\documentclass[letterpaper]{article} 
\usepackage{aaai2027}  
\usepackage[hyphens]{url}  
\usepackage{graphicx} 
\usepackage{natbib}  
\usepackage{caption} 
\usepackage{algorithm}
\usepackage{algorithmic}

\nocopyright

\usepackage{booktabs}
\usepackage{multirow}
\usepackage{graphicx}
\usepackage{booktabs}
\usepackage{multirow}
\usepackage{makecell}
\usepackage{tikz}
\usepackage{graphicx}
\newcommand{\avgimp}[2]{#1{\scriptsize$\uparrow$#2}}

\usepackage{amsmath}
\usepackage{amssymb}
\usepackage{booktabs}
\usepackage{tabularx}
\usepackage{array}

\usepackage{newfloat}
\usepackage{listings}
\DeclareCaptionStyle{ruled}{labelfont=normalfont,labelsep=colon,strut=off} 
\floatstyle{ruled}
\newfloat{listing}{tb}{lst}{}
\floatname{listing}{Listing}

\usepackage{booktabs}

\title{Benchmarking and Enhancing LLMs for Rule-Intensive Review of National Standard Documents}
\author{
    Tao Wang\textsuperscript{\rm 1}\equalcontrib,
    Qihao Yang\textsuperscript{\rm 2}\equalcontrib,
    Rongjiao Liang\textsuperscript{\rm 1},
    Lianghong Lin\textsuperscript{\rm 3},
    Haitao Wang\textsuperscript{\rm 3},
    Xinyu Cao\textsuperscript{\rm 3},
    Tianyong Hao\textsuperscript{\rm 1}\corresponding
}
\affiliations{
    \textsuperscript{\rm 1}School of Computer Science, South China Normal University, Guangzhou, China\\
    \textsuperscript{\rm 2}School of Artificial Intelligence, Shanghai Jiao Tong University, Shanghai, China\\
    \textsuperscript{\rm 3}China National Institute of Standardization, Beijing, China\\
}

\begin{document}

\maketitle

\begin{abstract}
Large language models (LLMs) are increasingly used to support complex professional tasks, yet their capabilities in rule-intensive document review remain insufficiently evaluated. National standard documents, such as China’s GB/T standards, provide a representative setting for such evaluation because they are lengthy, highly structured, and governed by explicit rules concerning scope, terminology, normative wording, and cross-section consistency. However, existing benchmarks primarily focus on domain knowledge and question answering, with limited attention to the intrinsic quality review of professional documents. Meanwhile, such reviews still rely heavily on human experts, making them costly and difficult to scale. To bridge this gap, we introduce \textbf{GB/T-Bench}, the first benchmark for the structured review of national standard documents. It consists of the GB/T Review Taxonomy, a hierarchical schema covering document structure, scope alignment, normative modality, terminology consistency, and normative references, further decomposed into 25 diagnosable error types. A controllable counterexample generation mechanism is proposed, combining deterministic rules with constrained LLM rewriting to process 488 documents and produce 7,306 traceable review error instances for evaluation. We also build a diagnosis-oriented evaluation protocol based on the exact matching of error location, review dimension, and error type, together with document-level coverage metrics. We further propose \textbf{GB/T-Reviewer}, a multi-agent review framework that converts review knowledge into specialized skills and coordinates global inspection, targeted diagnosis, rule scanning, and result verification. Experiments on 14 mainstream LLMs reveal a substantial human-LLM gap, with the strongest model achieving only 0.3280 CMCS compared with 0.6640 for experts. GB/T-Reviewer raises the best CMCS to 0.5094, demonstrating the value of structured skill coordination for rule-intensive document review. This work paves the way for trustworthy AI in standardization and other high-stakes document domains. 
\end{abstract}

\section{Introduction}



Large language models (LLMs) are increasingly used to support professional tasks involving long documents, domain knowledge, and complex reasoning \cite{zhao-etal-2024-docmath}, yet their capabilities in rule-intensive document review remain insufficiently evaluated. Such tasks require models to understand content, apply explicit rules, locate defects, distinguish closely related error types, and verify cross-section consistency \cite{deng-etal-2025-longdocurl}. Whether general-purpose LLMs can support precise and accountable professional review therefore remains an open question.

\begin{figure}[t]
    \centering
    \includegraphics[width=\columnwidth]{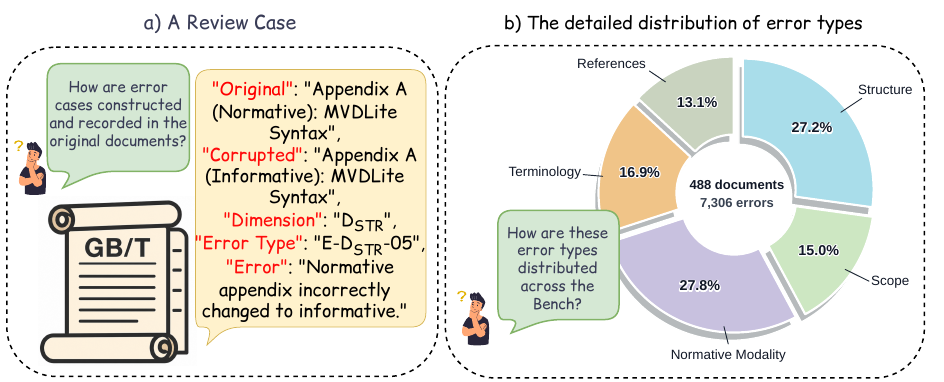}
    \caption{A GB/T document review case and dataset statistics of GB/T-Bench.}
    \label{fig:tease}
\end{figure}

National standard documents, including China’s GB/T standards, provide a representative setting for this problem. They encode technical requirements that guide product design, quality control, industrial practice, and public governance. Unlike ordinary technical texts, standard documents follow detailed rules for document structure, scope, normative wording, terminology, and references. Reviewers must determine, for example, whether the main body exceeds the declared scope, whether terms are used consistently, and whether expressions such as “shall,” “should,” and “may” convey the intended level of obligation. With more than 40,000 national standards currently in force in China, this review process still relies heavily on expert labor, resulting in high costs, long review cycles, and limited scalability.

Existing research covers only adjacent aspects of this task \cite{zhang-etal-2024-marathon}. Domain-specific benchmarks such as LegalBench \cite{guha2023legalbench}, LawBench \cite{fei2024lawbench}, PubMedQA \cite{jin2019pubmedqa}, and EduBench \cite{xu2026edubench} mainly assess whether LLMs can understand and apply professional knowledge. Studies on regulatory and standards-related texts, including Polisis \cite{harkous2018polisis}, PolicyQA \cite{ahmad2020policyqa}, SPaR.txt \cite{kruiper2021spar}, and CODE-ACCORD \cite{hettiarachchi2025code}, typically treat regulations as valid inputs and examine whether external artifacts comply with them. Reviewing a standard document itself reverses this setting: the rule-bearing text becomes the object of inspection, and the model must identify where the document violates its own drafting rules or internal constraints. This requires document-grounded diagnosis rather than answer-level correctness, including precise localization, error classification, and cross-section verification. Current benchmarks and general-purpose agent frameworks provide limited support for evaluating these coupled capabilities.

In this paper, we study an important yet underexplored problem: \textit{how to systematically evaluate and improve LLMs for reviewing the intrinsic quality of national standard documents?} Addressing this problem involves two major technical challenges. The first challenge is to establish a diagnostic evaluation foundation. Standard review involves heterogeneous errors across document structure, clause semantics, and cross-section consistency, while naturally occurring errors are costly to collect and annotate. Evaluation must therefore formalize review targets, construct controllable and traceable test cases, and assess whether models correctly locate and classify each defect. The second challenge is to make LLM-based review reliable under heterogeneous rule constraints. A single model must combine global document understanding with fine-grained rule checking and cross-section verification, which often results in missed or imprecise diagnoses. This calls for decomposing review knowledge into specialized skills and coordinating global inspection, targeted analysis, rule-based checking, and result verification \cite{zhao-etal-2024-longagent, 10.1145/3711896.3736570}.

To address the first challenge, we introduce GB/T-Bench, which integrates the GB/T Review Taxonomy, a controllable counterexample generation mechanism, and a diagnosis-oriented evaluation protocol. The taxonomy covers document structure, scope alignment, normative modality, terminology consistency, and normative references, further divided into 25 diagnosable error types. Combining deterministic rules with constrained LLM rewriting, the generation mechanism processes 488 GB/T documents and produces 7,306 traceable review error instances. Figure~\ref{fig:tease} illustrates an example error instance and the statistics of GB/T-Bench. The evaluation protocol requires predictions to match the error location, review category, and error type. To address the second challenge, we propose GB/T-Reviewer, a multi-agent framework that organizes review knowledge into specialized inspection skills. It coordinates global review, category specialists, error-type agents, and rule-based scanners to collect complementary diagnoses, which are then validated, merged, and deduplicated before final prediction. During evaluation, GB/T-Reviewer receives only the document under review and remains independent of the counterexample construction process.

We evaluate 14 closed-source and open-source LLMs on GB/T-Bench. The results show that current models are more effective at detecting explicit structural and scope errors, but remain limited in normative judgment, terminology alignment, and cross-section verification. The strongest LLM achieves only 0.3280 CMCS, while GB/T-Reviewer raises the best score to 0.5094 and improves most evaluated models. These results suggest that structured skill decomposition and coordinated verification can improve LLM-based review of rule-intensive professional documents.

The main contributions of this paper are as follows:

\begin{itemize}

    \item The GB/T-Bench is proposed, the first benchmark for evaluating LLMs in national standard document review. It establishes a GB/T review taxonomy covering 5 review dimensions and 25 fine-grained error types, together with a controllable benchmark and diagnosis-oriented evaluation protocol.


    \item The GB/T-Reviewer is proposed, a multi-agent review framework for national standard document review, which improves overall review performance and fine-grained diagnostic capability by coordinating specialized review modules.

    \item Extensive experiments are conducted on 14 mainstream large language models. The results systematically characterize their strengths and limitations and validate the effectiveness of the proposed agent-based enhancement framework.
\end{itemize}

\section{Related Work}


\paragraph{Benchmarks for Professional Document Understanding and Review.}
Existing benchmarks for professional documents have gradually expanded from domain knowledge and professional reasoning to document-level understanding. Comprehensive benchmarks such as LegalBench \cite{guha2023legalbench} and LexGLUE \cite{chalkidis2022lexglue} primarily evaluate reasoning and language understanding in specialized domains. Document-level benchmarks, including CUAD \cite{hendrycks2021cuad}, ContractNLI \cite{koreeda2021contractnli}, QASPER \cite{dasigi-etal-2021-dataset}, and ACORD \cite{wang-etal-2025-acord}, further focus on clause localization, evidence retrieval, document question answering, and hypothesis verification. Meanwhile, long-document benchmarks such as LongBench \cite{bai2024longbench}, MMLongBench-Doc \cite{ma2024mmlongbench}, and LongDocURL \cite{deng-etal-2025-longdocurl} assess information retrieval, content understanding, and long-context or cross-page reasoning. PeerRead \cite{kang2018dataset}, MAUD \cite{wang2023maud}, and FinMME \cite{luo2025finmme} respectively target acceptance prediction, specialized legal document understanding, and multimodal financial reasoning. However, existing benchmarks are largely built around predefined questions, designated text spans, or specific reasoning objectives, and thus remain insufficient for evaluating the open-ended quality review of complete, rule-governed professional documents. To address this gap, GB/T-Bench establishes a hierarchical review taxonomy grounded in GB/T drafting requirements and expert review practices, and uses it to guide the design of a diagnosis-level evaluation protocol.

\paragraph{Domain-Specific LLM Agents.}
LLM agents enhance the ability of language models to handle complex tasks by combining model reasoning with external knowledge \cite{yang2025llm, yang2025hskbenchmark}, tool invocation, and iterative verification. MRKL \cite{2022arXiv220500445K} integrates language models with knowledge sources, symbolic modules, and computational tools, while ReAct \cite{yao2023react} interleaves reasoning with interactions with the external environment. Toolformer \cite{schick2023toolformer} and ToolLLM \cite{qin2024toolllm} improve models' ability to select and invoke tools, whereas TaskWeaver \cite{qiao2023taskweaver}, AutoGen \cite{wu2024autogen}, and CRITIC \cite{gou2024critic} support code execution, multi-agent collaboration, and tool-feedback-based self-correction, respectively. However, existing general-purpose agent frameworks primarily focus on task planning, tool use, and collaborative reasoning, without explicitly modeling the rules and review requirements of professional documents, making them difficult to directly adapt to document review. To address this limitation, GB/T-Reviewer organizes standard-review knowledge into domain-specific review capabilities and collaborative workflows, enabling more systematic and reliable document diagnosis.

\begin{figure*}[t]
    \centering
    \includegraphics[width=\textwidth]{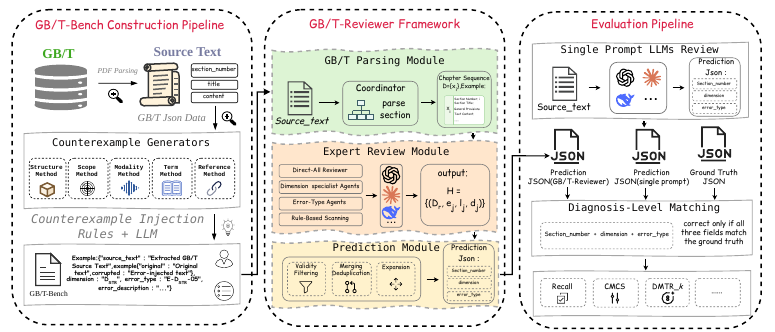}
\caption{Overview of GB/T-Bench and GB/T-Reviewer, consisting of a traceable error-injection-based dataset construction pipeline, a multi-agent GB/T-Reviewer framework for rule-intensive document review, and a systematic evaluation pipeline for assessing model predictions.}
    \label{fig:framework}
\end{figure*}

\section{GB/T-Bench}

This study investigates four aspects: the construction of the GB/T review taxonomy, dataset construction, benchmark evaluation design, and the development of the GB/T review agent framework. 
The overall workflow is shown in Figure \ref{fig:framework}.

\subsection{GB/T Review Taxonomy}

Reviewing national standard documents differs from answering predefined questions: an expert must systematically examine the complete document according to the relevant drafting requirements and identify its potential defects as comprehensively as possible. We define this process as Rule-Intensive Review, in which a complete standard document is provided as input to an LLM, which is required to produce structured diagnostic results containing the defect location, review dimension, fine-grained error type, and error description according to a predefined GB/T review taxonomy. Constructed based on GB/T drafting requirements and expert review practice, the taxonomy adopts a two-level structure consisting of five review dimensions and 25 review points. It specifies what should be reviewed and how defects should be categorized, while remaining independent of the procedure used to generate erroneous samples. We denote the five review dimensions as
\(\mathcal{D}_{\mathrm{rev}}=\{D_{\mathrm{STR}},D_{\mathrm{SCP}},D_{\mathrm{NMS}},D_{\mathrm{TER}},D_{\mathrm{NREF}}\}\),
corresponding to Structure Review, Scope Review, Normative Modality Strength Review, Terminology Review, and Normative Reference Review, respectively.
Each review dimension represents a core type of review behavior, while each review point defines a specific issue that can be independently identified, annotated, and evaluated.

Specifically, the review capabilities are organized at two levels.
First, the \textbf{document-level capabilities}, denoted as
\(\mathcal{D}_{\mathrm{doc}}=\{D_{\mathrm{STR}},D_{\mathrm{SCP}}\}\),
focus on the overall structure and content boundaries of standard documents.
The \(D_{\mathrm{STR}}\) dimension examines whether section organization, section order, title naming, and appendix attributes conform to GB/T 1.1, \textit{Directives for Standardization}.
The \(D_{\mathrm{SCP}}\) dimension verifies the consistency between the ``Scope'' section and the main body, preventing clauses from exceeding the declared applicable objects and boundaries of the standard.

Second, the \textbf{clause-level capabilities}, denoted as
\(\mathcal{D}_{\mathrm{clause}}=\{D_{\mathrm{NMS}},D_{\mathrm{TER}},D_{\mathrm{NREF}}\}\),
focus on normative expressions and semantic consistency within specific clauses.
The \(D_{\mathrm{NMS}}\) dimension evaluates whether normative modal verbs such as ``shall'', ``should'', and ``may'' are used with appropriate obligation strength.
The \(D_{\mathrm{TER}}\) dimension verifies whether term usage in the main body is consistent with the ``Terms and Definitions'' section.
The \(D_{\mathrm{NREF}}\) dimension checks whether the normative reference list is complete, accurate, and free from redundant references.

\subsection{GB/T-Bench Construction and Quality Control}
Based on the proposed GB/T review taxonomy, we construct
GB/T-Bench, a traceable benchmark dataset for structured
review of GB/T standard documents. To generate review
errors in a controllable manner, we propose a taxonomy-guided
counterexample generation mechanism. The mechanism adopts
a dimension-guided error-injection paradigm, in which
structural and formatting errors are generated through
deterministic rules, whereas semantic errors are generated
or rewritten by LLMs under rule-based constraints. Each instance preserves the original section, corrupted section, error location, review dimension, error type, and description, ensuring complete traceability. As shown in Table~\ref{tab:dataset_statistics}, the resulting dataset contains 64,991 sections and 7,306 error instances, covering five review dimensions and 25 fine-grained error types. Three trained graduate students perform end-to-end quality control throughout the error generation and injection process, removing candidate instances with mismatched labels, inaccurate locations, unintended modifications, duplication, or incomplete traceability.

\begin{table}[t]
\centering
\begin{tabular}{lr}
\toprule
\textbf{Item} & \textbf{Count} \\
\midrule
\#Unique Documents & 488 \\
\#Sections & 64,991 \\
\#Errors & 7,306 \\
Avg. Errors / Doc & 14.97 \\
Median Errors / Doc & 15 \\
Min / Max Errors per Doc & 10 / 18 \\
\#Dimensions & 5 \\
\#Error Types & 25 \\
\bottomrule
\end{tabular}
\caption{Statistics of the constructed GB/T-Bench.}
\label{tab:dataset_statistics}
\end{table}

To assess annotation reliability, we sample 30\% of the final dataset for human validation. Two domain experts and three additional trained graduate students independently evaluate the sampled instances using the same review taxonomy, after which the expert group and the graduate student group each establish consensus labels. The two groups achieve an exact agreement of 96.33\% on review dimensions, with a Cohen's $\kappa$ of 0.947. For fine-grained error types, the exact agreement is 95.92\%, with a Cohen's $\kappa$ of 0.955, while the exact agreement for the joint dimension-error-type labels is also 95.92\%. These results demonstrate that the proposed GB/T review taxonomy is readily operationalizable and supports stable and reliable annotation.

\section{GB/T-Reviewer}

We further propose GB/T-Reviewer, a multi-agent review framework designed to improve the performance of LLMs on national standard document review tasks. As shown in Figure \ref{fig:framework}, the GB/T-Reviewer consists of a GB/T parsing module, an expert review module, and a prediction module.





\paragraph{GB/T Parsing Module.}
The GB/T parsing module acts as the coordinator.
Given an input standard document \(S\), it parses the document into a section sequence
\(X=\{x_i\}_{i=1}^{n}\), where each section \(x_i=(p_i,h_i,c_i)\) contains a section number \(p_i\), a section title \(h_i\), and section content \(c_i\).
This module preserves the hierarchical structure and location information of the standard document, providing a unified context representation for subsequent review.
Through this process, model judgments can be grounded in specific section locations rather than remaining as vague full-document descriptions.


\paragraph{Expert Review Module.}
The expert review module performs multi-granularity analysis on the input standard document \(S\) and its section sequence \(X\), and generates a multi-source candidate error set \(H\).
Each candidate error is aligned with the review dimension and error type in the GB/T review taxonomy and can be represented as
\((D_r,e_j,l_j,d_j)\),
where \(D_r \in \mathcal{D}_{\mathrm{rev}}\) denotes the review dimension, \(e_j \in \mathcal{E}_r\) denotes the error type under that dimension, \(l_j\) denotes the error location, and \(d_j\) denotes the error description.

Specifically, this module organizes review agents at different granularities around the review dimensions, including the Direct-All Reviewer, Dimension Specialist Agents, and Error-Type Agents.
The Direct-All Reviewer performs broad-coverage scanning over the full document and all review dimensions to identify potential errors.
The Dimension Specialist Agents independently analyze different review dimensions to improve dimension-level coverage.
The Error-Type Agents further focus on high-risk review points under each review dimension and conduct targeted searches for specific error types.
In addition, the expert review module incorporates rule-based scanning mechanisms, using computable cues such as modal verbs, terminology indexes, section numbering, and reference relations to supplement candidates, thereby enhancing the detection of rule-based errors and cross-section consistency errors.
Finally, this module outputs the candidate set
\(H=\{(D_r,e_j,l_j,d_j)\}\),
which is then passed to the prediction module for filtering, fusion, and standardization.

\begin{table*}[t]
\centering
\resizebox{\textwidth}{!}{%
\begin{tabular}{llccccc}
\toprule
\textbf{Category}
& \textbf{Method}
& \textbf{DMTR\_8}
& \textbf{DMTR\_9}
& \textbf{DMTR\_10}
& \textbf{Recall}
& \textbf{CMCS} \\

\midrule

\multirow{1}{*}{%
  \makecell[c]{\small\textbf{Human Review}}}

& human
& \textbf{0.9442}
& \textbf{0.9021}
& \textbf{0.8479}
& \textbf{0.8308}
& \textbf{0.6640} \\

\midrule

\multirow{10}{*}{%
  \makecell[c]{\small\textbf{closed-source}\\
               \small\textbf{LLMs review}}}
& GPT-5.6-sol
& \textbf{0.5435}
& \underline{0.3696}
& 0.2174
& \textbf{0.5203}
& \textbf{0.3280} \\

& GPT-5.5
& \underline{0.5410}
& \textbf{0.3934}
& \underline{0.2439}
& \underline{0.5052}
& \underline{0.3185} \\

& GPT-5.4
& 0.3873
& 0.2500
& 0.1189
& 0.4287
& 0.2577 \\

& GPT-5.4-mini
& 0.1025
& 0.0389
& 0.0143
& 0.3052
& 0.1615 \\

& GPT-5.3
& 0.1844
& 0.0881
& 0.0266
& 0.3620
& 0.1996 \\

& Claude-Opus-4.8
& 0.2357
& 0.1434
& 0.0717
& 0.2904
& 0.1730 \\

& Claude-Fable-5
& 0.4783
& 0.3261
& \textbf{0.2609}
& 0.4993
& 0.3165 \\

& Gemini-3.5-Flash
& 0.3012
& 0.1721
& 0.0820
& 0.4087
& 0.2403 \\

& Qwen3.7-Plus
& 0.3443
& 0.1885
& 0.0943
& 0.4351
& 0.2583 \\

& \textit{Average}
& 0.3465
& 0.2189
& 0.1256
& 0.4172
& 0.2504 \\

\midrule

\multirow{6}{*}{%
  \makecell[c]{\small\textbf{open-source}\\
               \small\textbf{LLMs review}}}
& DeepSeek-v4-Pro
& \textbf{0.2193}
& \textbf{0.1230}
& \textbf{0.0492}
& \textbf{0.3961}
& \textbf{0.2269} \\

& DeepSeek-v4-Flash
& 0.1455
& 0.0738
& 0.0307
& \underline{0.3456}
& \underline{0.1910} \\

& MiniMax-M2.7
& 0.0656
& 0.0287
& 0.0102
& 0.2516
& 0.1296 \\

& Qwen2.5-VL-72B-Instruct
& 0.0820
& 0.0246
& 0.0082
& 0.2731
& 0.1431 \\

& Qwen3-235B-A22B-Instruct-2507
& \underline{0.1496}
& \underline{0.0820}
& \underline{0.0471}
& 0.2551
& 0.1449 \\

& \textit{Average}
& 0.1324
& 0.0664
& 0.0291
& 0.3043
& 0.1671 \\

\midrule

\multirow{10}{*}{%
  \makecell[c]{\small\textbf{GB/T-Reviewer with}\\
               \small\textbf{closed-source LLMs}}}
& GPT-5.6-sol
& \avgimp{0.8718}{0.3283}
& \avgimp{\underline{0.7919}}{0.4223}
& \avgimp{0.5385}{0.3211}
& \avgimp{0.6906}{0.1703}
& \avgimp{0.4697}{0.1417} \\

& GPT-5.5
& \avgimp{\underline{0.9026}}{0.3616}
& \avgimp{\textbf{0.8445}}{0.4511}
& \avgimp{\textbf{0.7860}}{0.5421}
& \avgimp{\textbf{0.7314}}{0.2262}
& \avgimp{\textbf{0.5094}}{0.1909} \\

& GPT-5.4
& \avgimp{0.8315}{0.4442}
& \avgimp{0.7656}{0.5156}
& \avgimp{\underline{0.6754}}{0.5565}
& \avgimp{0.6642}{0.2355}
& \avgimp{0.4457}{0.1880} \\

& GPT-5.4-mini
& \avgimp{0.4778}{0.3753}
& \avgimp{0.4073}{0.3684}
& \avgimp{0.3116}{0.2973}
& \avgimp{0.5221}{0.2169}
& \avgimp{0.3182}{0.1567} \\

& GPT-5.3
& \avgimp{0.8457}{0.6613}
& \avgimp{0.5658}{0.4777}
& \avgimp{0.3435}{0.3169}
& \avgimp{0.6014}{0.2394}
& \avgimp{0.3758}{0.1762} \\

& Claude-Opus-4.8
& \avgimp{0.7234}{0.4877}
& \avgimp{0.4681}{0.3247}
& \avgimp{0.4213}{0.3496}
& \avgimp{0.5694}{0.2790}
& \avgimp{0.3814}{0.2084} \\

& Claude-Fable-5
& \avgimp{\textbf{0.9137}}{0.4354}
& \avgimp{0.7778}{0.4517}
& \avgimp{0.6111}{0.3502}
& \avgimp{\underline{0.7008}}{0.2015}
& \avgimp{\underline{0.4757}}{0.1592} \\

& Gemini-3.5-Flash
& \avgimp{0.5324}{0.2312}
& \avgimp{0.3525}{0.1804}
& \avgimp{0.2852}{0.2032}
& \avgimp{0.5296}{0.1209}
& \avgimp{0.3205}{0.0802} \\

& Qwen3.7-Plus
& \avgimp{0.8289}{0.4846}
& \avgimp{0.6726}{0.4841}
& \avgimp{0.5991}{0.5048}
& \avgimp{0.6704}{0.2353}
& \avgimp{0.4473}{0.1890} \\

& \textit{Average}
& \avgimp{0.7698}{0.4233}
& \avgimp{0.6273}{0.4084}
& \avgimp{0.5080}{0.3824}
& \avgimp{0.6311}{0.2139}
& \avgimp{0.4160}{0.1656} \\

\midrule

\multirow{6}{*}{%
  \makecell[c]{\small\textbf{GB/T-Reviewer with}\\
               \small\textbf{open-source LLMs}}}
& DeepSeek-v4-Pro
& \avgimp{\underline{0.6654}}{0.4461}
& \avgimp{\underline{0.4866}}{0.3636}
& \avgimp{\underline{0.3799}}{0.3307}
& \avgimp{\underline{0.5647}}{0.1686}
& \avgimp{\underline{0.3535}}{0.1266} \\

& DeepSeek-v4-Flash
& \avgimp{\textbf{0.7730}}{0.6275}
& \avgimp{\textbf{0.6428}}{0.5690}
& \avgimp{\textbf{0.3942}}{0.3635}
& \avgimp{\textbf{0.5894}}{0.2438}
& \avgimp{\textbf{0.3624}}{0.1714} \\

& MiniMax-M2.7
& \avgimp{0.3720}{0.3064}
& \avgimp{0.3001}{0.2714}
& \avgimp{0.1558}{0.1456}
& \avgimp{0.4660}{0.2144}
& \avgimp{0.2632}{0.1336} \\

& Qwen2.5-VL-72B-Instruct
& \avgimp{0.5410}{0.4590}
& \avgimp{0.3764}{0.3518}
& \avgimp{0.2196}{0.2114}
& \avgimp{0.4921}{0.2190}
& \avgimp{0.2822}{0.1391} \\

& Qwen3-235B-A22B-Instruct-2507
& \avgimp{0.5484}{0.3988}
& \avgimp{0.3710}{0.2890}
& \avgimp{0.2640}{0.2169}
& \avgimp{0.5258}{0.2707}
& \avgimp{0.3317}{0.1868} \\

& \textit{Average}
& \avgimp{0.5800}{0.4476}
& \avgimp{0.4354}{0.3690}
& \avgimp{0.2827}{0.2536}
& \avgimp{0.5276}{0.2233}
& \avgimp{0.3186}{0.1515} \\

\bottomrule
\end{tabular}%
}
\caption{Performance comparison of five review settings on GB/T-Bench. The smaller-font values below the main GB/T-Reviewer scores indicate absolute changes relative to the corresponding backbone model review results. Higher values indicate better performance; bold and underlined values denote the best and second-best results in each category, respectively. Recall refers to Diagnosis Recall throughout the table.}
\label{tab:performance_comparison}
\end{table*}

\paragraph{Prediction Module.}
Given the candidate error set \(H\) generated by multiple agents and rule-based scanners, the prediction module first validates the predicted review dimension \(D_r\) and error type \(e_j\) against the GB/T review taxonomy. It then integrates the heterogeneous candidates through confidence-based fusion and deduplication. For ambiguous predictions, the module further refines error locations and explores neighboring error types to reduce omissions caused by cross-section dependencies and ambiguous category boundaries. Finally, it outputs structured review results containing the review dimension, error type, document location, and error description.

\section{Experiments and Results}

\subsection{Implementation Details}


\paragraph{Baselines.} We compare the proposed method under five types of
baselines:
(1) Human Review, including two expert reviewers and three trained graduate students as human performance references;
(2) closed-source LLMs review;
(3) open-source LLMs review;
(4) GB/T-Reviewer with closed-source LLMs;
(5) GB/T-Reviewer with open-source LLMs.

\paragraph{Models.}
The evaluated closed-source models include GPT-5.6-sol, GPT-5.5, GPT-5.4, GPT-5.4-mini, GPT-5.3, Gemini-3.5-Flash, Claude-Opus-4.8, Claude-Fable-5 and Qwen3.7-Plus \cite{qwen2.5-1m}. 
In addition, we evaluate five open-source models: DeepSeek-v4-Pro \cite{xu2026DeepSeek}, DeepSeek-v4-Flash \cite{xu2026DeepSeek}, MiniMax-M2.7 \cite{chen2026minimax}, Qwen2.5-VL-72B-Instruct \cite{team2023qwen, wang2024qwen2, wu2025qwen} and Qwen3-235B-A22B-Instruct-2507 \cite{qwen2.5-1m}.

\paragraph{Evaluation Metrics.}
We report DMTR\_8--DMTR\_10, Recall, and CMCS as the main evaluation metrics. All metrics adopt exact diagnosis-level matching: a prediction is considered correct only when its section position, review dimension, and error type all match the ground-truth annotation.

\noindent\textbf{Recall.}
It measures the proportion of ground-truth errors correctly diagnosed across the dataset:
\[
\mathrm{Recall}
=
\frac{\sum_{i=1}^{N} T_i}
{\sum_{i=1}^{N} |G_i|},
\]
where \(G_i\) denotes the ground-truth error set of document \(i\), \(T_i\) denotes the number of correctly matched diagnoses, and \(N\) denotes the total number of documents.

\noindent\textbf{DMTR\_\textit{k}.}
It measures the proportion of documents in which at least \(k\) errors are correctly diagnosed:
\[
\mathrm{DMTR\_}k
=
\frac{1}{N}
\sum_{i=1}^{N}
\mathbb{I}(T_i \ge k),
\]
where \(\mathbb{I}(\cdot)\) denotes the indicator function. We report DMTR\_8, DMTR\_9, and DMTR\_10 to assess high-coverage diagnostic performance.

\noindent\textbf{CMCS.}
It provides an overall assessment of diagnostic performance by jointly considering error coverage and penalties for missed and redundant predictions:
\[
\mathrm{CMCS}
=
\frac{
\sum_{i=1}^{N}
|G_i|
\left(
\frac{T_i}{|G_i|}
e^{-\lambda \frac{|G_i|-T_i}{|G_i|}}
\left(
1-\alpha\frac{F_i}{F_i+\beta}
\right)
\right)
}{
\sum_{i=1}^{N}|G_i|
},
\]
where \(P_i\) denotes the predicted error set of document \(i\), \(F_i=\max(|P_i|-T_i,0)\) denotes the number of redundant predictions, and \(\lambda\), \(\alpha\), and \(\beta\) control the missed-error and redundant-prediction penalties. We set \(\lambda=1.0\), \(\alpha=0.15\), and \(\beta=100.0\) in all experiments.
\subsection{Main Results}

Table~\ref{tab:performance_comparison} summarizes the performance of different model settings, where Average denotes the mean performance across all valid models under each setting. The following analysis focuses on two questions: (1)  \textit{Do current LLMs still exhibit a substantial capability gap compared with human experts in rule-intensive standard document review?} (2)  \textit{Can GB/T-Reviewer effectively improve the review capability of LLMs for this task?}

\textbf{For the first question}, \textit{current LLMs still exhibit a substantial capability gap compared with human experts in rule-intensive standard document review.} Across both closed-source and open-source LLM settings, even the best-performing models remain significantly below human performance. For example, GPT-5.6-sol achieves a CMCS of only 0.3280, approximately half of the Human Review score (0.6640), while its Recall decreases from 0.8308 to 0.5203. Moreover, under the strictest diagnostic criterion (DMTR\_10), GPT-5.6-sol achieves only 0.2174, compared with 0.8479 for Human Review. Similarly, the best-performing open-source model, DeepSeek-v4-Pro, achieves a CMCS of only 0.2269 and a DMTR\_10 score of merely 0.0492, further highlighting the considerable gap between current LLMs and expert-level review under complex rule constraints. These results indicate that although current LLMs are capable of identifying some document defects, they remain inadequate in accurate error localization and fine-grained diagnostic reasoning across entire documents. To ensure a fair comparison, Human Review was conducted on a separate, non-overlapping 50\% subset randomly sampled from the portion of the final GB/T-Bench not used for annotation quality control, using the same GB/T review taxonomy and evaluation protocol as the model evaluation and without access to the gold annotations. The trained graduate students participating in Human Review were independent of those involved in dataset quality validation, thereby avoiding potential bias introduced during dataset construction.

\textbf{For the second question}, \textit{GB/T-Reviewer effectively improves the performance of LLMs on rule-intensive standard document review through structured multi-agent collaboration.} After incorporating GB/T-Reviewer, all backbone models consistently achieve better performance, demonstrating the effectiveness of structured multi-agent collaboration in enhancing document review capability. For closed-source LLMs, GB/T-Reviewer improves the average CMCS from 0.2504 under the Single Prompt setting to 0.4160. Among them, the GPT-5.5-based GB/T-Reviewer achieves the highest CMCS of 0.5094, representing an absolute improvement of 0.1909 over its corresponding Single Prompt baseline. Meanwhile, its DMTR\_10 score increases from 0.2439 to 0.7860, indicating substantial improvements not only in error coverage but also in fine-grained diagnostic capability. Similar gains are observed for open-source models, where the average CMCS increases from 0.1671 to 0.3186. Overall, these results demonstrate that GB/T-Reviewer consistently enhances both the overall review performance and the fine-grained diagnostic capability of LLMs for rule-intensive standard document review.

\begin{table}[t]
\centering
\small
\resizebox{\columnwidth}{!}{%
\begin{tabular}{lccccc}
\toprule
\textbf{Model} & $\mathbf{D_{\mathrm{STR}}}$ & $\mathbf{D_{\mathrm{SCP}}}$ & $\mathbf{D_{\mathrm{NMS}}}$ & $\mathbf{D_{\mathrm{TER}}}$ & $\mathbf{D_{\mathrm{NREF}}}$ \\
\midrule
GPT-5.6-sol & \textbf{0.7581} & 0.7596 & 0.4150 & 0.2294 & 0.3407 \\
Claude-Fable-5 & 0.6508 & 0.7087 & 0.4162 & 0.2477 & 0.4286 \\
GPT-5.5 & 0.7194 & \textbf{0.8181} & \textbf{0.4166} & 0.1909 & 0.4389 \\
GPT-5.4 & 0.5587 & 0.6965 & 0.3261 & 0.1772 & 0.3960 \\
GPT-5.4-mini & 0.3199 & 0.4333 & 0.1697 & 0.2540 & 0.4828 \\
GPT-5.3 & 0.4826 & 0.5430 & 0.2597 & 0.2015 & 0.3302 \\
Claude-Opus-4.8 & 0.3557 & 0.4196 & 0.2184 & 0.1319 & 0.3657 \\
DeepSeek-v4-Pro & 0.5033 & 0.6033 & 0.3114 & 0.1820 & 0.3939 \\
DeepSeek-v4-Flash & 0.4181 & 0.5338 & 0.2445 & 0.1626 & 0.4316 \\
Gemini-3.5-Flash & 0.4987 & 0.6088 & 0.3158 & \textbf{0.2597} & 0.3835 \\
Qwen3.7-Plus & 0.5441 & 0.6700 & 0.3384 & 0.2168 & 0.4284 \\
MiniMax-M2.7 & 0.2922 & 0.3346 & 0.2164 & 0.1197 & 0.3177 \\
Qwen2.5-VL-72B-Instruct & 0.2534 & 0.3839 & 0.1323 & 0.1820 & \textbf{0.6040} \\
Qwen3-235B-A22B-Instruct-2507 & 0.2474 & 0.3611 & 0.1589 & 0.1934 & 0.4347 \\
\bottomrule
\end{tabular}%
}
\caption{Dimension-wise Diagnosis Recall of the evaluated LLMs. Higher values indicate better performance; bold values denote the best result in each review dimension.}
\label{tab:single_prompt_dimension_recall}
\end{table}


\subsection{Hierarchical Diagnostic Capability Analysis}

To further analyze the capability bottlenecks of LLMs in GB/T standard document review, this paper decomposes the review process into three increasingly strict levels: Location Recall, Dimension Recall, and Diagnosis Recall.
Location Recall measures whether a model can locate the correct section containing an error; Dimension Recall further requires the model to identify the correct review dimension; and Diagnosis Recall is the strictest metric, requiring the section location, review dimension, and error type to all match the ground-truth annotation.

\begin{figure}[t]
    \centering
    \includegraphics[width=\columnwidth]{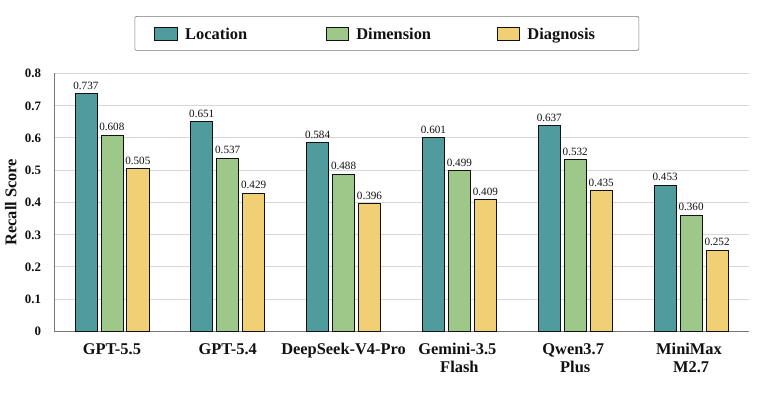}
    \caption{Location, dimension, and diagnosis recall across six models.}
    \label{fig:recall_three_levels}
\end{figure}

For clarity in the main text, this paper selects six representative models for analysis, namely GPT-5.5, GPT-5.4, DeepSeek-v4-Pro, Gemini-3.5-Flash, Qwen3.7-Plus, and MiniMax-M2.7.
These models cover strong, medium-performing, and relatively weak systems, allowing a more balanced comparison across different capability levels.
As shown in Figure~\ref{fig:recall_three_levels}, all six models exhibit a consistent decreasing trend from Location Recall to Dimension Recall and then to Diagnosis Recall.
This indicates that the models are relatively more capable of locating suspicious sections but still face clear difficulties in assigning the correct review dimension and identifying fine-grained error types.

\subsection{Dimension-wise Capability Boundaries of LLMs}

\begin{figure}[t]
    \centering
    \includegraphics[width=\columnwidth]{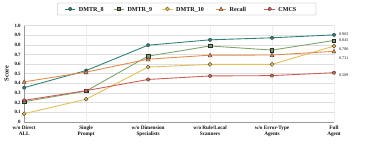}
    \caption{Ablation study results of GPT-5.5. Recall refers to Diagnosis Recall.}
    \label{fig:ablation}
\end{figure}

As shown in Table~\ref{tab:single_prompt_dimension_recall}, current LLMs exhibit clear capability differences across review dimensions. Models perform relatively well on \(D_{\mathrm{STR}}\) (document structure review) and \(D_{\mathrm{SCP}}\) (scope consistency review), where errors are primarily associated with explicit document organization, including section hierarchies, heading structures, and scope statements. Consequently, these tasks mainly rely on global document understanding and recognition of explicit structural patterns, allowing existing LLMs to achieve relatively stable performance.

In contrast, performance drops substantially on \(D_{\mathrm{NMS}}\) (normative modality strength review) and \(D_{\mathrm{TER}}\) (terminology consistency review). Unlike document-level review, these tasks require models to determine whether normative expressions convey the intended level of obligation and whether terminology remains consistent with its formal definitions throughout the document. Such judgments depend not only on semantic understanding, but also on compliance with drafting rules, definition consistency verification, and rule-constrained reasoning. Similarly, \(D_{\mathrm{NREF}}\) (normative reference review) requires cross-section evidence integration to verify the consistency between in-text citations and the normative reference list. Although some models can detect explicit citation errors, they remain unstable in more fine-grained cases, such as redundant references and confusion between normative and informative references.

Overall, current LLMs are effective at identifying document errors with explicit structural cues, but remain limited in rule-intensive reasoning involving normative judgment, definition consistency, and cross-section verification. This limitation motivates GB/T-Reviewer to decompose complex review tasks into specialized modules.

\subsection{Ablation Study}

To further investigate how each component of GB/T-Reviewer addresses the capability limitations identified above, we conduct an ablation study using GPT-5.5 as the backbone model. As shown in Figure~\ref{fig:ablation}, the complete GB/T-Reviewer consistently achieves the best performance across all evaluation metrics, demonstrating that the proposed modules provide complementary reasoning capabilities rather than redundant improvements.

Among all ablation settings, removing Direct ALL results in the largest performance degradation. Compared with the complete model, CMCS decreases from 0.5094 to approximately 0.22, Recall drops to approximately 0.42, and DMTR\_10 decreases to approximately 0.08, even performing worse than the Single Prompt baseline. This finding indicates that Direct ALL establishes a unified global understanding of the entire document before specialist reasoning begins. Without this holistic context, downstream agents operate on fragmented evidence, substantially reducing both error coverage and diagnostic consistency.

Removing Dimension Specialists also leads to a notable performance decline, with DMTR\_10 decreasing from 0.7860 to approximately 0.57 and CMCS decreasing to approximately 0.44. This suggests that decomposing heterogeneous review criteria into dimension-specific reasoning effectively reduces interference among different review tasks, thereby improving fine-grained diagnostic capability. Similarly, removing Error-Type Agents reduces DMTR\_10 and CMCS to approximately 0.60 and 0.48, respectively, indicating that explicitly modeling high-risk error categories is essential for distinguishing semantically similar but normatively different defects.

Furthermore, removing Rule/Local Scanners consistently decreases Recall, CMCS, and DMTR\_10, demonstrating that deterministic rule matching and localized contextual evidence effectively complement semantic reasoning by recovering errors that may otherwise be overlooked and by improving error matching and diagnostic accuracy.

The ablation results show that the components are complementary: Direct ALL provides global context, Dimension Specialists and Error-Type Agents strengthen dimension-level reasoning and fine-grained diagnosis, respectively, while Rule/Local Scanners introduce deterministic evidence, jointly improving the overall performance of GB/T-Reviewer.


\section{Conclusion}

This paper introduces GB/T-Bench, a benchmark for evaluating LLM capabilities in national standard document review, together with a review taxonomy and a controllable, traceable dataset constructed through multidimensional error injection. Comprehensive evaluations reveal that existing LLMs remain limited in rule-intensive document review. To address these limitations, we propose GB/T-Reviewer, a multi-agent framework that organizes review knowledge into specialized skills and coordinates complementary review processes. Experimental results show that GB/T-Reviewer consistently improves performance across different backbone models and narrows the gap between LLMs and expert reviewers. GB/T-Bench further provides a foundation for reliable AI-assisted review in structured, rule-governed, and other high-stakes professional document domains.







\bibliography{aaai2027}

\clearpage
\appendix

\section{Appendix}
The appendix provides supplementary materials for the main paper in five sections: Detailed GB/T Review Taxonomy, Detailed Explanation of Evaluation Metrics, Supplementary Results of Hierarchical Diagnostic Capability Analysis, Token Cost Analysis, and Prompts. These materials provide further details on the review taxonomy, metric definitions, hierarchical diagnostic results, token consumption, and the prompt and rule configurations used by the review components.


\section{A.Detailed GB/T Review Taxonomy}
\label{sec:detailed_taxonomy}

This section provides a detailed description of the GB/T Review Taxonomy adopted in this study. As shown in Table~\ref{tab:review_dimensions}, the taxonomy comprises five review dimensions:
\[
\mathcal{D}_{\mathrm{rev}}
=
\{D_{\mathrm{STR}},D_{\mathrm{SCP}},D_{\mathrm{NMS}},
D_{\mathrm{TER}},D_{\mathrm{NREF}}\}.
\]
These dimensions capture the core quality requirements of GB/T documents at both the document and clause levels.

Specifically, \(D_{\mathrm{STR}}\) and \(D_{\mathrm{SCP}}\) represent document-level review dimensions. \(D_{\mathrm{STR}}\) examines whether document structure, numbering, formatting, and annex organization conform to GB/T~1.1, whereas \(D_{\mathrm{SCP}}\) assesses whether the main content is consistent with the declared scope. The remaining dimensions operate primarily at the clause level. \(D_{\mathrm{NMS}}\) evaluates the correctness of normative modal expressions, \(D_{\mathrm{TER}}\) examines the consistency between terms, definitions, and their usage, and \(D_{\mathrm{NREF}}\) assesses the completeness, necessity, and ordering of normative references.

For fine-grained diagnosis, each review dimension is further decomposed into concrete error types. Tables~\ref{tab:dstr_errors}--\ref{tab:dnref_errors} present the complete taxonomy of error types across the five dimensions. This hierarchical taxonomy enables GB/T-Bench to evaluate not only whether an error is detected, but also whether its review dimension and fine-grained error type are correctly identified. It therefore supports both overall performance evaluation and detailed diagnostic analysis.

\section{B. Detailed Explanation of Evaluation Metrics}

To evaluate the performance of different methods, we use five complementary metrics:
DMTR\_8, DMTR\_9, DMTR\_10, Recall, and CMCS. These metrics are designed for a
multi-error diagnosis task, where a model is expected not only to determine whether a
document contains errors, but also to recover as many ground-truth error items as
possible with the correct location, review dimension, and error type. In the main
paper, we focus on DMTR\_8, DMTR\_9, and DMTR\_10 because they reflect high-threshold
diagnostic completion. To provide a more fine-grained view of model behavior under
different diagnostic thresholds, we further report the complete DMTR\_1--DMTR\_10
results in Table~\ref{tab:dmtr_1_to_10}.

For each document, the ground-truth annotations and model predictions are represented
as item lists. Each item is matched at the diagnosis level using the tuple:
\[
(\text{section\_number}, \text{dimension}, \text{error\_type}).
\]
A predicted item is counted as correct only when these fields match a ground-truth
item. To avoid over-counting duplicated predictions, the implementation uses
multiset-based matching: for each unique diagnosis tuple, the number of matches is
the minimum of its ground-truth count and prediction count.

\paragraph{DMTR\_k.}
DMTR stands for Diagnosis Match Threshold Rate. It measures the proportion of
documents in which the model correctly diagnoses at least \(k\) ground-truth error
items. For document \(i\), let \(h_i\) denote the number of correctly matched
diagnosis items. Then:
\[
\mathrm{DMTR}\_k = \frac{1}{N}\sum_{i=1}^{N} \mathbb{I}(h_i \ge k),
\quad k \in \{1,\ldots,10\},
\]
where \(N\) is the number of evaluated documents and \(\mathbb{I}(\cdot)\) is the
indicator function.

DMTR is used because the task requires high-coverage diagnosis rather than merely
identifying one or two obvious errors. Lower thresholds such as DMTR\_1--DMTR\_7
show whether a model can recover a small or moderate number of errors, while higher
thresholds such as DMTR\_8, DMTR\_9, and DMTR\_10 impose stricter completeness
requirements. In particular, DMTR\_10 reflects whether the model can approach
near-complete diagnosis for documents with many annotated errors. Therefore,
Table~\ref{tab:dmtr_1_to_10} provides a more detailed threshold-wise analysis,
showing how model performance changes from relatively easy diagnostic targets to
high-completeness diagnostic targets.

\paragraph{Recall.}
Recall measures the overall item-level diagnosis coverage across the entire dataset.
Let \(H=\sum_i h_i\) be the total number of correctly matched diagnosis items and
\(G=\sum_i |G_i|\) be the total number of ground-truth items. The diagnosis recall is:
\[
\mathrm{Recall} = \frac{H}{G}.
\]
This metric captures how many ground-truth error items are successfully recovered by
the model. In this evaluation, a recall hit requires the correct section, dimension,
and error type, so it reflects diagnosis-level recovery rather than loose semantic
overlap.

\paragraph{CMCS.}
CMCS is a completeness-aware and noise-penalized score. It is based on a per-document
MCS score and then applies a soft penalty for excessive predictions. For document
\(i\), let \(G_i\) be the number of ground-truth items, \(h_i\) the number of correctly
matched diagnosis items, and \(P_i\) the number of predicted items. The per-document
recall is:
\[
r_i = \frac{h_i}{G_i}.
\]
The miss rate is:
\[
m_i = \frac{G_i - h_i}{G_i}.
\]
The MCS score is computed as:
\[
\mathrm{MCS}_i = r_i \cdot \exp(-\lambda m_i),
\]
where \(\lambda=1.0\). To discourage over-generation, the implementation further
defines a false-positive proxy:
\[
\mathrm{FP}_i = \max(P_i - h_i, 0),
\]
and a soft noise penalty:
\[
\mathrm{Penalty}_i = 1 - \alpha \cdot
\frac{\mathrm{FP}_i}{\mathrm{FP}_i+\beta},
\]
where \(\alpha=0.15\) and \(\beta=100.0\). The final CMCS is the ground-truth-weighted
average:
\[
\mathrm{CMCS} =
\frac{\sum_i G_i \cdot \mathrm{MCS}_i \cdot \mathrm{Penalty}_i}
{\sum_i G_i}.
\]

Overall, these metrics evaluate the task from complementary perspectives. DMTR
measures document-level diagnostic completion under different thresholds, Recall
measures corpus-level item recovery, and CMCS measures balanced diagnosis quality by
combining coverage, missed-error penalty, and soft noise control. The additional
DMTR\_1--DMTR\_10 results in Table~\ref{tab:dmtr_1_to_10} further reveal the full
performance trajectory of each model across increasingly strict diagnostic thresholds.


\section{C.Supplementary Results of Hierarchical Diagnostic Capability Analysis}
\label{sec:supp_hierarchical_analysis}

\begin{figure*}[t]
    \centering
    \includegraphics[width=\textwidth]{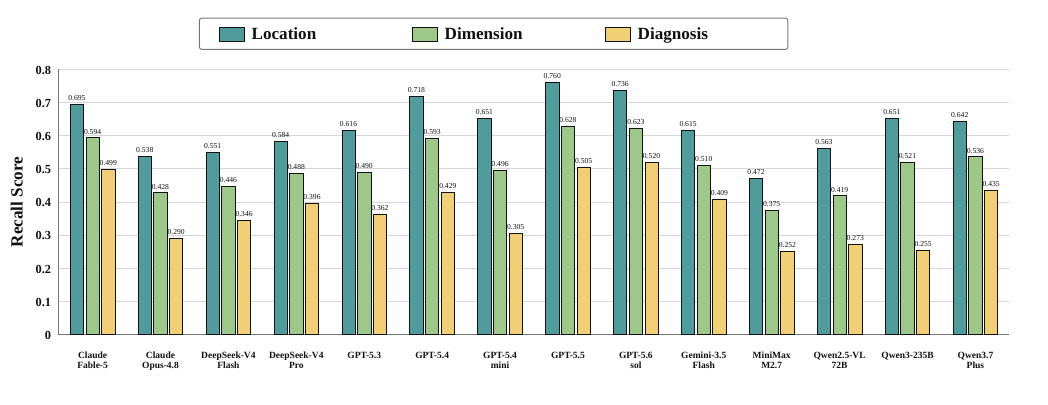}
\caption{Hierarchical recall of 14 standalone LLMs on GB/T-Dataset. Location, Dimension, and Diagnosis represent increasingly strict matching criteria, requiring the correct error location, review dimension, and fine-grained error type, respectively.}
    \label{fig:hierarchical_recall}
\end{figure*}

This section presents the complete hierarchical diagnostic results across the evaluated standalone LLMs. As shown in Figure~\ref{fig:hierarchical_recall}, model performance is compared at three progressively stricter levels: error location, review dimension, and fine-grained diagnosis. The supplementary results further provide dimension-level comparisons and error-type-level observations, revealing how diagnostic performance degrades as the evaluation criterion shifts from error localization to precise error-type identification.

\begin{table*}[t]
\centering
\small

\begin{tabularx}{\textwidth}{@{}>{\centering\arraybackslash}p{0.16\textwidth}>{\centering\arraybackslash}p{0.16\textwidth}>{\centering\arraybackslash}p{0.18\textwidth}X@{}}
\toprule
\textbf{ID}
& \textbf{Level}
& \textbf{Dimension}
& \textbf{Review Task} \\
\midrule

\(D_{\mathrm{STR}}\)
& Document-level
& Structure Review
& Determine whether the chapter structure conforms to GB/T~1.1 requirements. \\

\(D_{\mathrm{SCP}}\)
& Document-level
& Scope Review
& Determine whether the body content is consistent with the stated scope. \\

\(D_{\mathrm{NMS}}\)
& Clause-level
& Normative Modality Review
& Determine whether modal expressions such as ``shall,'' ``should,'' and ``may'' are used correctly. \\

\(D_{\mathrm{TER}}\)
& Clause-level
& Terminology Review
& Determine whether terms are used consistently with their definitions. \\

\(D_{\mathrm{NREF}}\)
& Clause-level
& Normative Reference Review
& Determine whether the normative reference list is complete and non-redundant. \\

\bottomrule
\end{tabularx}

\caption{Review dimensions in the GB/T Review Taxonomy.}
\label{tab:review_dimensions}
\end{table*}

\begin{table*}[t]
\centering
\small

\begin{tabularx}{\textwidth}{@{}>{\centering\arraybackslash}p{0.20\textwidth}>{\centering\arraybackslash}p{0.25\textwidth}X@{}}
\toprule
\textbf{Error Code}
& \textbf{Error Type}
& \textbf{Description} \\
\midrule

\texttt{E-D\_STR-01}
& Incorrect heading numbering format
& The chapter numbering format does not comply with the specification, e.g., using ``Chapter 1'' instead of ``1,'' or using ``1)'' instead of ``1.1.'' \\

\texttt{E-D\_STR-02}
& Non-standard font or font size
& The font or font size used in the body text or headings does not comply with Appendix D of GB/T~1.1, e.g., using Times New Roman instead of Songti for Chinese text. \\

\texttt{E-D\_STR-03}
& Incorrect list format
& The list marker is incorrectly used, e.g., using ``\textbullet'' instead of an em dash, or failing to use the 1), 2) format for lettered list items. \\

\texttt{E-D\_STR-04}
& Incorrect note/example format
& Notes or examples are not placed on independent lines, or their indentation does not meet the formatting requirements. \\

\texttt{E-D\_STR-05}
& Incorrect header/footer format
& The header does not include the standard number, or the page number is placed incorrectly. \\

\texttt{E-D\_STR-06}
& Incorrect appendix label
& The appendix is not marked as normative or informative, or the appendix numbering does not start with the letter A. \\

\bottomrule
\end{tabularx}

\caption{Error types under the structure review dimension \(D_{\mathrm{STR}}\).}
\label{tab:dstr_errors}
\end{table*}

\begin{table*}[t]
\centering
\small

\begin{tabularx}{\textwidth}{@{}>{\centering\arraybackslash}p{0.20\textwidth}>{\centering\arraybackslash}p{0.25\textwidth}X@{}}
\toprule
\textbf{Error Code}
& \textbf{Error Type}
& \textbf{Description} \\
\midrule

\texttt{E-D\_SCP-01}
& Body content exceeds the defined scope
& The body specifies product types, use scenarios, or technical parameters that are not mentioned in the scope section. \\

\texttt{E-D\_SCP-02}
& Body content fails to cover scope commitments
& The scope states that the document applies to a certain type of product, but the body contains no corresponding technical requirements. \\

\texttt{E-D\_SCP-03}
& Requirement clauses included in the scope
& The scope section contains requirement expressions such as ``shall'' or ``should,'' violating the principle that the scope should contain only descriptive statements. \\

\bottomrule
\end{tabularx}

\caption{Error types under the scope review dimension \(D_{\mathrm{SCP}}\).}
\label{tab:dscp_errors}
\end{table*}

\begin{table*}[t]
\centering
\small

\begin{tabularx}{\textwidth}{@{}>{\centering\arraybackslash}p{0.20\textwidth}>{\centering\arraybackslash}p{0.25\textwidth}X@{}}
\toprule
\textbf{Error Code}
& \textbf{Error Type}
& \textbf{Description} \\
\midrule

\texttt{E-D\_NMS-01}
& ``Should'' used instead of ``shall''
& A mandatory requirement is expressed as a recommendation, weakening the binding force of the clause. \\

\texttt{E-D\_NMS-02}
& ``Shall'' used instead of ``should''
& A recommendation is expressed as a mandatory requirement, resulting in excessive constraint. \\

\texttt{E-D\_NMS-03}
& ``May'' used instead of ``shall''
& A mandatory requirement is expressed as permission, causing the clause to lose its binding force. \\

\texttt{E-D\_NMS-04}
& Mixed use of ``shall'' and ``must''
& ``Must'' is used instead of ``shall,'' whereas GB/T~1.1 does not adopt ``must'' as the normative auxiliary for requirements. \\

\texttt{E-D\_NMS-05}
& Use of vague modal expressions
& Vague expressions such as ``as far as possible,'' ``preferably,'' or ``generally'' are used instead of standardized modal auxiliaries. \\

\texttt{E-D\_NMS-06}
& ``Should not'' used instead of ``shall not''
& A prohibition is expressed as a non-recommendation, weakening the prohibitive force of the clause. \\

\texttt{E-D\_NMS-07}
& Inconsistency between modality and clause content
& The clause content clearly expresses a mandatory safety requirement, but it is formulated using ``should'' or ``may.'' \\

\bottomrule
\end{tabularx}

\caption{Error types under the normative modality review dimension \(D_{\mathrm{NMS}}\).}
\label{tab:dnms_errors}
\end{table*}

\begin{table*}[t]
\centering
\small

\begin{tabularx}{\textwidth}{@{}>{\centering\arraybackslash}p{0.20\textwidth}>{\centering\arraybackslash}p{0.25\textwidth}X@{}}
\toprule
\textbf{Error Code}
& \textbf{Error Type}
& \textbf{Description} \\
\midrule

\texttt{E-D\_TER-01}
& Use of undefined terms
& A technical term appears in the main body, but no corresponding definition is provided in the Terms and Definitions section. \\

\texttt{E-D\_TER-02}
& Defined term not used in the main body
& A term is defined in the Terms and Definitions section, but it is never used in the main body. \\

\texttt{E-D\_TER-03}
& Inconsistent use of term and definition
& A term is used in the main body with a conceptual meaning inconsistent with its formal definition. \\

\texttt{E-D\_TER-04}
& Multiple terms used for the same concept
& Multiple near-synonymous terms are used for the same technical concept across different sections. \\

\texttt{E-D\_TER-05}
& Discontinuous term numbering
& The numbering of terms is discontinuous, such as 3.1, 3.2, and 3.4, with 3.3 missing. \\

\bottomrule
\end{tabularx}

\caption{Error types under the terminology review dimension \(D_{\mathrm{TER}}\).}
\label{tab:dter_errors}
\end{table*}

\begin{table*}[t]
\centering
\small

\begin{tabularx}{\textwidth}{@{}>{\centering\arraybackslash}p{0.20\textwidth}>{\centering\arraybackslash}p{0.28\textwidth}X@{}}
\toprule
\textbf{Error Code}
& \textbf{Error Type}
& \textbf{Description} \\
\midrule

\texttt{E-D\_NREF-01}
& Missing normative reference
& A standard is normatively cited in the body, but it is not listed in the Normative References section. \\

\texttt{E-D\_NREF-02}
& Redundant normative reference
& A standard is listed in the Normative References section, but it is never cited in the body. \\

\texttt{E-D\_NREF-03}
& Informative reference included in normative list
& A standard mentioned only in a note or example is included in the Normative References section. \\

\texttt{E-D\_NREF-04}
& Incorrect ordering of references
& Normative references are not arranged in the required order of standard numbers. \\

\bottomrule
\end{tabularx}

\caption{Error types under the normative reference review dimension \(D_{\mathrm{NREF}}\).}
\label{tab:dnref_errors}
\end{table*}

\renewcommand{\avgimp}[2]{#1{\tiny\(\uparrow\!#2\)}}
\begin{table*}[t]
\centering

\scriptsize
\setlength{\tabcolsep}{2pt}

\resizebox{\linewidth}{!}{%
\begin{tabular}{llcccccccccc}
\toprule
\textbf{Setting}
& \textbf{Model}
& \textbf{DMTR\_1}
& \textbf{DMTR\_2}
& \textbf{DMTR\_3}
& \textbf{DMTR\_4}
& \textbf{DMTR\_5}
& \textbf{DMTR\_6}
& \textbf{DMTR\_7}
& \textbf{DMTR\_8}
& \textbf{DMTR\_9}
& \textbf{DMTR\_10} \\
\midrule


\multirow{14}{*}{%
\makecell[c]{\textbf{Direct}\\
             \textbf{LLM Review}}}

& GPT-5.5
& 0.9959
& 0.9959
& 0.9857
& 0.9713
& 0.9303
& 0.8422
& 0.7275
& 0.5410
& 0.3934
& 0.2439 \\

& GPT-5.6-sol
& 1.0000
& 1.0000
& 1.0000
& 0.9783
& 0.9348
& 0.8043
& 0.6957
& 0.5435
& 0.3696
& 0.2174 \\

& GPT-5.4
& 0.9057
& 0.9037
& 0.8934
& 0.8586
& 0.7807
& 0.6803
& 0.5656
& 0.3873
& 0.2500
& 0.1189 \\

& GPT-5.4-mini
& 0.9016
& 0.8852
& 0.8053
& 0.6947
& 0.5410
& 0.3689
& 0.2111
& 0.1025
& 0.0389
& 0.0143 \\

& GPT-5.3
& 0.9488
& 0.9406
& 0.8914
& 0.8238
& 0.6762
& 0.5020
& 0.3299
& 0.1844
& 0.0881
& 0.0266 \\

& Claude-Opus-4.8
& 0.7446
& 0.7399
& 0.7351
& 0.6945
& 0.6348
& 0.5465
& 0.3914
& 0.2357
& 0.1434
& 0.0717 \\

& Claude-Fable-5
& 1.0000
& 1.0000
& 1.0000
& 0.9348
& 0.8913
& 0.7391
& 0.6304
& 0.4783
& 0.3261
& 0.2609 \\

& Deepseek-v4-Pro
& 0.9795
& 0.9693
& 0.9160
& 0.8525
& 0.7316
& 0.6291
& 0.4344
& 0.2193
& 0.1230
& 0.0492 \\

& Deepseek-v4-Flash
& 0.9385
& 0.9119
& 0.8709
& 0.7828
& 0.6516
& 0.4570
& 0.3012
& 0.1455
& 0.0738
& 0.0307 \\

& Gemini-3.5-Flash
& 0.9727
& 0.9643
& 0.9370
& 0.8824
& 0.7899
& 0.6387
& 0.4664
& 0.3012
& 0.1721
& 0.0820 \\

& Qwen3.7-Plus
& 0.9836
& 0.9754
& 0.9508
& 0.9037
& 0.8197
& 0.6721
& 0.5123
& 0.3443
& 0.1885
& 0.0943 \\

& MiniMax-M2.7
& 0.8545
& 0.7971
& 0.6844
& 0.5615
& 0.4016
& 0.2357
& 0.1250
& 0.0656
& 0.0287
& 0.0102 \\

& Qwen2.5-VL-72B-Instruct
& 0.8668
& 0.8279
& 0.7520
& 0.6025
& 0.4529
& 0.2971
& 0.1721
& 0.0820
& 0.0246
& 0.0082 \\

& Qwen3-235B-A22B-Instruct-2507
& 0.6414
& 0.6250
& 0.5922
& 0.5492
& 0.4713
& 0.3832
& 0.2541
& 0.1496
& 0.0820
& 0.0471 \\

\midrule


\multirow{14}{*}{%
\makecell[c]{\textbf{GB/T-Reviewer}\\
             \textbf{with LLMs}}}

& GPT-5.5
& \avgimp{1.0000}{0.0041}
& \avgimp{1.0000}{0.0041}
& \avgimp{1.0000}{0.0143}
& \avgimp{1.0000}{0.0287}
& \avgimp{1.0000}{0.0697}
& \avgimp{1.0000}{0.1578}
& \avgimp{0.9867}{0.2592}
& \avgimp{0.9026}{0.3616}
& \avgimp{0.8445}{0.4511}
& \avgimp{0.7860}{0.5421} \\

& GPT-5.6-sol
& 1.0000
& 1.0000
& 1.0000
& \avgimp{1.0000}{0.0217}
& \avgimp{1.0000}{0.0652}
& \avgimp{0.9744}{0.1701}
& \avgimp{0.9744}{0.2787}
& \avgimp{0.8718}{0.3283}
& \avgimp{0.7919}{0.4223}
& \avgimp{0.5385}{0.3211} \\

& GPT-5.4
& \avgimp{1.0000}{0.0943}
& \avgimp{1.0000}{0.0963}
& \avgimp{1.0000}{0.1066}
& \avgimp{1.0000}{0.1414}
& \avgimp{1.0000}{0.2193}
& \avgimp{0.9867}{0.3064}
& \avgimp{0.9600}{0.3944}
& \avgimp{0.8315}{0.4442}
& \avgimp{0.7656}{0.5156}
& \avgimp{0.6754}{0.5565} \\

& GPT-5.4-mini
& \avgimp{1.0000}{0.0984}
& \avgimp{1.0000}{0.1148}
& \avgimp{1.0000}{0.1947}
& \avgimp{0.9867}{0.2920}
& \avgimp{0.9600}{0.4190}
& \avgimp{0.8667}{0.4978}
& \avgimp{0.7067}{0.4956}
& \avgimp{0.4778}{0.3753}
& \avgimp{0.4073}{0.3684}
& \avgimp{0.3116}{0.2973} \\

& GPT-5.3
& \avgimp{1.0000}{0.0512}
& \avgimp{1.0000}{0.0594}
& \avgimp{1.0000}{0.1086}
& \avgimp{1.0000}{0.1762}
& \avgimp{0.9545}{0.2783}
& \avgimp{0.9091}{0.4071}
& \avgimp{0.8636}{0.5337}
& \avgimp{0.8457}{0.6613}
& \avgimp{0.5658}{0.4777}
& \avgimp{0.3435}{0.3169} \\

& Claude-Opus-4.8
& \avgimp{0.9362}{0.1916}
& \avgimp{0.9362}{0.1963}
& \avgimp{0.8723}{0.1372}
& \avgimp{0.8723}{0.1778}
& \avgimp{0.8298}{0.1950}
& \avgimp{0.7872}{0.2407}
& \avgimp{0.7872}{0.3958}
& \avgimp{0.7234}{0.4877}
& \avgimp{0.4681}{0.3247}
& \avgimp{0.4213}{0.3496} \\

& Claude-Fable-5
& 1.0000
& 1.0000
& 1.0000
& \avgimp{1.0000}{0.0652}
& \avgimp{1.0000}{0.1087}
& \avgimp{1.0000}{0.2609}
& \avgimp{1.0000}{0.3696}
& \avgimp{0.9137}{0.4354}
& \avgimp{0.7778}{0.4517}
& \avgimp{0.6111}{0.3502} \\

& Deepseek-v4-Pro
& \avgimp{1.0000}{0.0205}
& \avgimp{1.0000}{0.0307}
& \avgimp{0.9787}{0.0627}
& \avgimp{0.9787}{0.1262}
& \avgimp{0.9362}{0.2046}
& \avgimp{0.8936}{0.2645}
& \avgimp{0.8085}{0.3741}
& \avgimp{0.6654}{0.4461}
& \avgimp{0.4866}{0.3636}
& \avgimp{0.3799}{0.3307} \\

& Deepseek-v4-Flash
& \avgimp{1.0000}{0.0615}
& \avgimp{1.0000}{0.0881}
& \avgimp{1.0000}{0.1291}
& \avgimp{1.0000}{0.2172}
& \avgimp{0.9867}{0.3351}
& \avgimp{0.8933}{0.4363}
& \avgimp{0.8400}{0.5388}
& \avgimp{0.7730}{0.6275}
& \avgimp{0.6428}{0.5690}
& \avgimp{0.3942}{0.3635} \\

& Gemini-3.5-Flash
& \avgimp{1.0000}{0.0273}
& \avgimp{0.9996}{0.0353}
& \avgimp{0.9939}{0.0569}
& \avgimp{0.9888}{0.1064}
& \avgimp{0.9715}{0.1816}
& \avgimp{0.9384}{0.2997}
& \avgimp{0.8821}{0.4157}
& \avgimp{0.5324}{0.2312}
& \avgimp{0.3525}{0.1804}
& \avgimp{0.2852}{0.2032} \\

& Qwen3.7-Plus
& \avgimp{1.0000}{0.0164}
& \avgimp{1.0000}{0.0246}
& \avgimp{1.0000}{0.0492}
& \avgimp{1.0000}{0.0963}
& \avgimp{1.0000}{0.1803}
& \avgimp{1.0000}{0.3279}
& \avgimp{0.9565}{0.4442}
& \avgimp{0.8289}{0.4846}
& \avgimp{0.6726}{0.4841}
& \avgimp{0.5991}{0.5048} \\

& MiniMax-M2.7
& \avgimp{1.0000}{0.1455}
& \avgimp{0.9857}{0.1886}
& \avgimp{0.9857}{0.3013}
& \avgimp{0.9571}{0.3956}
& \avgimp{0.9000}{0.4984}
& \avgimp{0.6857}{0.4500}
& \avgimp{0.5571}{0.4321}
& \avgimp{0.3720}{0.3064}
& \avgimp{0.3001}{0.2714}
& \avgimp{0.1558}{0.1456} \\

& Qwen2.5-VL-72B-Instruct
& \avgimp{1.0000}{0.1332}
& \avgimp{1.0000}{0.1721}
& \avgimp{1.0000}{0.2480}
& \avgimp{1.0000}{0.3975}
& \avgimp{0.9516}{0.4987}
& \avgimp{0.8710}{0.5739}
& \avgimp{0.7097}{0.5376}
& \avgimp{0.5410}{0.4590}
& \avgimp{0.3764}{0.3518}
& \avgimp{0.2196}{0.2114} \\

& Qwen3-235B-A22B-Instruct-2507
& \avgimp{1.0000}{0.3586}
& \avgimp{1.0000}{0.3750}
& \avgimp{1.0000}{0.4078}
& \avgimp{1.0000}{0.4508}
& \avgimp{0.9516}{0.4803}
& \avgimp{0.9194}{0.5362}
& \avgimp{0.8710}{0.6169}
& \avgimp{0.5484}{0.3988}
& \avgimp{0.3710}{0.2890}
& \avgimp{0.2640}{0.2169} \\

\bottomrule
\end{tabular}%
}
\caption{DMTR\_1--DMTR\_10 scores under direct LLM review and GB/T-Reviewer settings. The smaller values accompanying the GB/T-Reviewer scores indicate absolute changes relative to the corresponding backbone models.}
\label{tab:dmtr_1_to_10}
\end{table*}

\begin{figure*}[t]
    \centering
    \includegraphics[width=\textwidth]{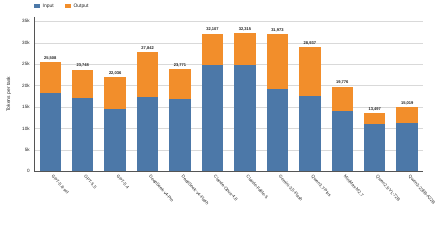}
\caption{Average per-task input and output token consumption across the evaluated LLMs. Each stacked bar shows the input and output token usage, with the total number of tokens reported above the bar.}
\label{fig:token_cost_analysis}
\end{figure*}


\begin{table*}[t]
\centering
\footnotesize

\begin{minipage}{\textwidth}
\hrule
\smallskip

\begin{minipage}[t]{0.485\textwidth}
\raggedright

\textbf{[Role]:}\par
You are an expert reviewer of GB/T national standard
documents. Your task is to identify defects in a standard
document according to GB/T~1.1.

\medskip
\textbf{[Instruction]:}\par
Read the complete GB/T standard document and review all
five dimensions.

\medskip
\textbf{[Review Dimensions]:}\par

\texttt{D\_STR}: \textbf{Structure Review.}
Check missing sections, incorrect ordering, non-standard
section names, confusion between the preface and introduction,
incorrect appendix classification, excessive hierarchy depth,
and related structural defects.

\smallskip
\texttt{D\_SCP}: \textbf{Scope Review.}
Check whether the scope is consistent with the objects,
applicability boundaries, document title, and scope statements
described in the main body.

\smallskip
\texttt{D\_NMS}: \textbf{Normative Modality Strength Review.}
Check whether modal expressions such as ``shall,'' ``should,''
``may,'' ``shall not,'' and ``should not'' express the intended
level of obligation and are consistent with the clause content.

\smallskip
\texttt{D\_TER}: \textbf{Terminology Review.}
Check undefined terms, unused defined terms, inconsistent term
usage, multiple terms for the same concept, and discontinuous
term numbering.

\smallskip
\texttt{D\_NREF}: \textbf{Normative Reference Review.}
Check missing, redundant, incorrectly classified, or incorrectly
ordered normative references.

\medskip
\textbf{[Error Types]:}\par

\textbf{\texttt{D\_STR} (Structure Review):}\par
\texttt{E-D\_STR-01}: Missing mandatory element;\par
\texttt{E-D\_STR-02}: Incorrect section order;\par
\texttt{E-D\_STR-03}: Non-standard section name;\par
\texttt{E-D\_STR-04}: Confusion between the preface and
introduction;\par
\texttt{E-D\_STR-05}: Incorrect appendix classification;\par
\texttt{E-D\_STR-06}: Excessive hierarchy depth.

\smallskip
\textbf{\texttt{D\_SCP} (Scope Review):}\par
\texttt{E-D\_SCP-01}: Main body exceeds the declared scope;\par
\texttt{E-D\_SCP-02}: Main body fails to cover a declared scope
commitment;\par
\texttt{E-D\_SCP-03}: The scope contains normative requirements.

\textbf{\texttt{D\_NMS} (Normative Modality Strength Review):}\par
\texttt{E-D\_NMS-01}: ``should'' used instead of ``shall'';\par
\texttt{E-D\_NMS-02}: ``shall'' used instead of ``should'';\par
\texttt{E-D\_NMS-03}: ``may'' used instead of ``shall'';\par
\texttt{E-D\_NMS-04}: Inconsistent use of ``shall'' and
``must'';\par
\texttt{E-D\_NMS-05}: Use of vague modal expressions;\par
\texttt{E-D\_NMS-06}: ``should not'' used instead of
``shall not'';\par
\texttt{E-D\_NMS-07}: Contradiction between the modal expression
and the clause content.

\end{minipage}
\hfill
\begin{minipage}[t]{0.485\textwidth}
\raggedright

\smallskip
\textbf{\texttt{D\_TER} (Terminology Review):}\par
\texttt{E-D\_TER-01}: Undefined term used in the main body;\par
\texttt{E-D\_TER-02}: Defined term not used in the main body;\par
\texttt{E-D\_TER-03}: Term usage inconsistent with its
definition;\par
\texttt{E-D\_TER-04}: Multiple terms used for the same concept;\par
\texttt{E-D\_TER-05}: Discontinuous term numbering.

\smallskip
\textbf{\texttt{D\_NREF} (Normative Reference Review):}\par
\texttt{E-D\_NREF-01}: Missing normative reference;\par
\texttt{E-D\_NREF-02}: Redundant normative reference;\par
\texttt{E-D\_NREF-03}: Informative reference mixed into the
normative reference list;\par
\texttt{E-D\_NREF-04}: Incorrect ordering of normative references.

\medskip
\textbf{[Review Requirements]:}\par

1. Cover all five dimensions rather than focusing on only one
dimension.\par

2. Report every error supported by the document and report each
error only once.\par

3. If no error is found, return an empty array \texttt{[]}.\par

4. \texttt{section\_number} must identify the section containing
the error. Use \texttt{null} for an unnumbered section and an
appendix identifier such as \texttt{"Appendix A"} for an appendix.\par

5. \texttt{reason} must summarize the cause of the error in one
sentence.\par

6. \texttt{error\_description} must follow the format
\texttt{"Error: ...; Suggested revision: ...; Basis: ..."}.

\medskip
\textbf{[Output Format]:}\par

Return a JSON array in which each item contains
\texttt{error\_type}, \texttt{dimension},
\texttt{section\_number}, \texttt{reason}, and
\texttt{error\_description}.

Return only the JSON array and no additional explanation.

\medskip
\textbf{[JSON Schema]:}\par

{\ttfamily
[\par
\quad \{\par
\qquad "error\_type": "E-D\_STR-01",\par
\qquad "dimension": "D\_STR",\par
\qquad "section\_number": "1",\par
\qquad "reason": "...",\par
\qquad "error\_description":\par
\qquad\quad "Error: ...; Suggested revision: ...;\par
\qquad\quad Basis: ..."\par
\quad \}\par
]\par
}

\end{minipage}

\smallskip
\hrule
\end{minipage}

\caption{Shared prompt used in both the LLMs Review setting and the Direct-All Reviewer of GB/T-Reviewer.}
\label{tab:global_review_prompt}
\end{table*}

\begin{table*}[t]
\centering
\footnotesize

\begin{minipage}{\textwidth}
\hrule
\smallskip

\begin{minipage}[t]{0.485\textwidth}
\raggedright

\textbf{[Full-Scan User Prompt]}

\medskip
\textbf{[Locally Pre-filtered Context]:}\par
The following context was identified by local rule-based
pre-filtering. It highlights potentially relevant sections,
reference sets, or terminology clues. First inspect this
context, then use the complete document to verify the
evidence, section number, and error type.

\smallskip
\texttt{\{focused\_context\}}

\medskip
\textbf{[Complete Document under Review]:}\par
\texttt{\{source\_text\}}

\medskip
\textbf{[Task]:}\par
Independently perform a full review of
\texttt{\{dim\}}:

\smallskip
\(\bullet\) Do not only search for omissions and do not rely
on the result of the Direct-All Reviewer.\par

\(\bullet\) Output every candidate error that may be valid
under the current dimension.\par

\(\bullet\) If one section may contain multiple error types,
output separate candidates.\par

\(\bullet\) For cross-section issues, output the
\texttt{section\_number} most likely to be used as the error
location by a human annotator. Key comparison sections may
also be included.\par

\smallskip
If no candidate error is found, return \texttt{[]}.

\medskip
\textbf{[Dimension-Specific Guides]}

\medskip
\textbf{\texttt{D\_STR} Specialist:}\par
Review only \texttt{D\_STR}: structural and hierarchical
organization of the standard. Focus on the overall structure,
front matter, clause hierarchy, appendices, reference sections,
table of contents, section numbering, and the ownership of
structural units.

\smallskip
You may output only \texttt{E-D\_STR-01} through
\texttt{E-D\_STR-06}.

\medskip
\textbf{\texttt{D\_SCP} Specialist:}\par
Review only \texttt{D\_SCP}: the standard scope, applicable
objects, normative-content boundaries, and standard
attributes. Check whether the scope is too broad or too
narrow, whether applicable objects are consistent throughout
the document, whether supporting sections match the declared
scope, and whether normative and informative content are
confused.

\smallskip
You may output only \texttt{E-D\_SCP-01},
\texttt{E-D\_SCP-02}, and \texttt{E-D\_SCP-03}.

\end{minipage}
\hfill
\begin{minipage}[t]{0.485\textwidth}
\raggedright

\textbf{\texttt{D\_NMS} Specialist:}\par
Review only \texttt{D\_NMS}: normative wording and the
expression of requirements in standard clauses. Focus on
``shall,'' ``should,'' ``may,'' ``can,'' ``must,''
``shall not,'' ``should not,'' and vague expressions. Also
check whether the subject, action, condition, and result of a
requirement are complete and whether the modal strength
agrees with the clause content.

\smallskip
You may output only \texttt{E-D\_NMS-01} through
\texttt{E-D\_NMS-07}.

\medskip
\textbf{\texttt{D\_TER} Specialist:}\par
Review only \texttt{D\_TER}: terminology definitions, term
usage, and terminology consistency. Focus on undefined
terms, defined terms that are never used, disagreement
between a term and its definition, multiple terms for the same
concept, and discontinuous term numbering.

\smallskip
You may output only \texttt{E-D\_TER-01} through
\texttt{E-D\_TER-05}.

\medskip
\textbf{\texttt{D\_NREF} Specialist:}\par
Review only \texttt{D\_NREF}: normative reference documents,
reference lists, external-document citations, and reference
consistency. Focus on missing or redundant references,
confusion between normative and informative references,
and inconsistencies in reference identifiers, publication
years, titles, versions, citation properties, or ordering.

\smallskip
You may output only \texttt{E-D\_NREF-01} through
\texttt{E-D\_NREF-04}.

\end{minipage}

\smallskip
\hrule
\end{minipage}

\caption{Prompt specification for the five Dimension Specialist sub-agents in GB/T-Reviewer.}
\label{tab:dimension_specialist_prompt}
\end{table*}

\begin{table*}[t]
\centering
\footnotesize

\begin{minipage}{\textwidth}
\hrule
\smallskip

\begin{minipage}[t]{0.485\textwidth}
\raggedright

\textbf{[System Prompt]}

\medskip
You are a GB/T standard-review Error-Type Agent.
You are responsible for identifying one error type only:

\smallskip
\texttt{\{dim\} / \{error\_type\}}.

\medskip
\textbf{[Current Dimension Rules]:}\par
\texttt{\{dimension\_guide\}}

\medskip
\textbf{[Mandatory Constraints]:}\par

1. \texttt{dimension} must be exactly
\texttt{"\{dim\}"}.\par

2. \texttt{error\_type} must be exactly
\texttt{"\{error\_type\}"}.\par

3. Output a candidate only when the target error is supported
by verifiable evidence from the document.\par

4. Do not output other error types or issues belonging to
neighboring review dimensions.\par

5. \texttt{section\_number} must identify the primary error
location most likely to be annotated by a human reviewer.
For cross-section issues, include supporting comparison
sections in the \texttt{evidence} field.\par

6. Do not output duplicate candidates referring to the same
underlying error at the same primary location.

\medskip
\textbf{[Output Format]:}\par

Return a JSON array. Each item must contain:

\smallskip
\(\bullet\) \texttt{error\_type}\par
\(\bullet\) \texttt{dimension}\par
\(\bullet\) \texttt{section\_number}\par
\(\bullet\) \texttt{reason}\par
\(\bullet\) \texttt{error\_description}\par
\(\bullet\) \texttt{evidence}\par
\(\bullet\) \texttt{confidence}\par

\smallskip
Return only the JSON array and no additional explanation.

\end{minipage}
\hfill
\begin{minipage}[t]{0.485\textwidth}
\raggedright

\textbf{[User Prompt]}

\medskip
\textbf{[Locally Focused Context]:}\par
\texttt{\{focused\_context\}}

\medskip
\textbf{[Complete Document under Review]:}\par
\texttt{\{source\_text\}}

\medskip
\textbf{[Task]:}\par

Identify only:

\smallskip
\texttt{\{dim\} / \{error\_type\}}.

\smallskip
Use the locally focused context to locate potential evidence,
and use the complete document to verify the target error type,
primary location, and cross-section consistency.

\smallskip
Medium-confidence candidates may be included to improve
recall only when they are supported by explicit textual or
cross-section evidence. Do not output speculative candidates
based solely on general drafting expectations.

\smallskip
If no supported candidate is found, return \texttt{[]}.

\medskip
\textbf{[Instantiated Dimension--Error-Type Pairs]:}\par

\texttt{D\_NMS / E-D\_NMS-02}\par
\texttt{D\_NMS / E-D\_NMS-07}\par
\texttt{D\_TER / E-D\_TER-01}\par
\texttt{D\_TER / E-D\_TER-02}\par
\texttt{D\_TER / E-D\_TER-03}\par
\texttt{D\_TER / E-D\_TER-04}\par
\texttt{D\_NREF / E-D\_NREF-01}

\medskip
These agents target selected high-risk error types that require
fine-grained semantic judgment or cross-section verification
and are frequently missed by broader review agents.

\end{minipage}

\smallskip
\hrule
\end{minipage}

\caption{Prompt specification for the Error-Type sub-agents
in GB/T-Reviewer.}
\label{tab:error_type_agent_prompt}
\end{table*}

\begin{table*}[t]
\centering
\footnotesize

\begin{minipage}{\textwidth}
\hrule
\smallskip

\begin{minipage}[t]{0.485\textwidth}
\raggedright

\textbf{[\texttt{D\_NMS} Sentence-Level Scanner]}

\medskip
\textbf{[Local Pre-filtering]:}\par
The local program extracts sentences containing modal
expressions corresponding to ``shall not,'' ``should not,''
``must,'' ``shall,'' ``should,'' ``may,'' ``as far as
possible,'' ``preferably,'' ``generally,'' ``in principle,''
and ``when necessary.''

\medskip
\textbf{[System Prompt]:}\par

You are a GB/T \texttt{D\_NMS} normative-modality review
agent. You will receive sentences extracted by local rules
because they contain modal expressions.

\smallskip
For each sentence, determine whether it may contain a
\texttt{D\_NMS} error. Medium-confidence candidates are
allowed. If one sentence may correspond to multiple modality
error types, output separate candidates.

\smallskip
\texttt{error\_type} must be selected from:

\smallskip
\texttt{E-D\_NMS-01}, \texttt{E-D\_NMS-02},
\texttt{E-D\_NMS-03}, \texttt{E-D\_NMS-04},
\texttt{E-D\_NMS-05}, \texttt{E-D\_NMS-06}, and
\texttt{E-D\_NMS-07}.

\smallskip
Return a JSON array. Each item must contain:

\smallskip
\(\bullet\) \texttt{error\_type}\par
\(\bullet\) \texttt{dimension}\par
\(\bullet\) \texttt{section\_number}\par
\(\bullet\) \texttt{reason}\par
\(\bullet\) \texttt{error\_description}\par
\(\bullet\) \texttt{evidence}\par
\(\bullet\) \texttt{confidence}\par

\smallskip
\texttt{dimension} must be exactly \texttt{"D\_NMS"}.
Return only the JSON array and no explanation.

\medskip
\textbf{[User Prompt Example]:}\par

{\ttfamily
[\par
\quad \{\par
\qquad "section\_number": "5.2",\par
\qquad "title": "Test Conditions",\par
\qquad "sentence": "The equipment shall operate\par
\qquad\quad under the specified conditions.",\par
\qquad "modal\_words": ["shall"]\par
\quad \}\par
]\par
}

\medskip
\textbf{[Generation Parameters]:}\par
\texttt{temperature = 0.0}\par
\texttt{max\_tokens = 7000}

\end{minipage}
\hfill
\begin{minipage}[t]{0.485\textwidth}
\raggedright

\textbf{[\texttt{D\_TER} Terminology-Index Scanner]}

\medskip
\textbf{[Local Pre-filtering]:}\par
The local program constructs a terminology index containing:

\smallskip
\(\bullet\) the terminology table;\par
\(\bullet\) the location of each definition;\par
\(\bullet\) locations where each term is used in the main body;\par
\(\bullet\) the number of term occurrences;\par
\(\bullet\) the term-number sequence; and\par
\(\bullet\) excerpts from technical clauses.

\medskip
\textbf{[System Prompt]:}\par

You are a GB/T \texttt{D\_TER} terminology-review agent.
You will receive a locally extracted terminology table, a
summary of term usage in the main body, and excerpts from
technical clauses.

\smallskip
Use the terminology index to identify \texttt{D\_TER} errors.
Focus on:

\smallskip
\texttt{E-D\_TER-01}: an undefined term is used in the
main body;\par
\texttt{E-D\_TER-02}: a defined term is never used in the
main body;\par
\texttt{E-D\_TER-03}: term usage is inconsistent with the
definition;\par
\texttt{E-D\_TER-04}: multiple terms are used for the same
concept;\par
\texttt{E-D\_TER-05}: term numbering is discontinuous.

\smallskip
Medium-confidence candidates are allowed, but every
candidate must include evidence.

\smallskip
Return a JSON array. Each item must contain
\texttt{error\_type}, \texttt{dimension},
\texttt{section\_number}, \texttt{reason},
\texttt{error\_description}, \texttt{evidence}, and
\texttt{confidence}.

\smallskip
\texttt{dimension} must be exactly \texttt{"D\_TER"}.
Return only the JSON array and no explanation.

\medskip
\textbf{[User Prompt]:}\par

\textbf{[Terminology Table]:}\par
\texttt{\{term\_entries\_json\}}

\smallskip
\textbf{[Main-Body Usage Summary]:}\par
\texttt{\{term\_usage\_summary\_json\}}

\smallskip
\textbf{[Technical-Clause Excerpts]:}\par
\texttt{\{technical\_element\_context\}}

\end{minipage}

\medskip
\textbf{[\texttt{D\_NREF} Reference-Consistency Scanner]}

\smallskip
The normative-reference scanner is primarily deterministic
and does not use a free-form LLM prompt. It compares the set
of standard identifiers cited in the main body, the set of
documents in the normative-reference list, the set of
documents in the bibliography or informative references,
and the ordering of entries in the normative-reference list.

\smallskip
It generates candidates according to the following rules:

\smallskip
\texttt{Reference cited in the main body but absent from the
normative-reference list}
\(\rightarrow\) \texttt{E-D\_NREF-01}.\par

\texttt{Reference listed but never cited in the main body}
\(\rightarrow\) \texttt{E-D\_NREF-02}.\par

\texttt{Reference mentioned only informatively but included
in the normative-reference list}
\(\rightarrow\) \texttt{E-D\_NREF-03}.\par

\texttt{Normative-reference entries are not in the required
order}
\(\rightarrow\) \texttt{E-D\_NREF-04}.

\smallskip
This deterministic component provides stable and
interpretable evidence without relying on free-form model
generation.

\smallskip
\hrule
\end{minipage}

\caption{Prompt specifications and deterministic rules for the Rule/Local Scanner sub-agents in GB/T-Reviewer.}
\label{tab:rule_local_scanner_prompts}
\end{table*}

\section{D.Token Cost Analysis}
\label{sec:token_cost_analysis}

Token consumption consists primarily of input and output tokens.

\textbf{Input Tokens.}
Input tokens are mainly contributed by the complete standard document, review rules, and task instructions. For most models, they account for the majority of the total token consumption.

\textbf{Output Tokens.}
Output tokens are primarily used to generate structured error predictions, diagnostic rationales, and error descriptions. Differences in response length therefore lead to variations in total token consumption across models.

\textbf{Total Token Consumption.}
Figure~\ref{fig:token_cost_analysis} summarizes the average input, output, and total token consumption per review task for each model. Overall, token usage varies substantially across models, and higher consumption does not necessarily correspond to better review performance. GPT-5.5 achieves strong diagnostic performance with a moderate token budget, indicating a favorable balance between efficiency and effectiveness.

\section{E.Prompts}
For reproducibility, we present the prompt specifications and rule configurations used by the review components in the following tables: the shared comprehensive-review prompt for the standalone LLMs Review setting and the Direct-All Reviewer in Table~\ref{tab:global_review_prompt}; the prompts for the five Dimension Specialist sub-agents in Table~\ref{tab:dimension_specialist_prompt}; the system and user prompts for the Error-Type sub-agents in Table~\ref{tab:error_type_agent_prompt}; and the prompts and deterministic rules for the Rule/Local Scanner sub-agents in Table~\ref{tab:rule_local_scanner_prompts}.

\end{document}